\documentclass[10pt,twocolumn,letterpaper]{article}
\pdfoutput=1
\usepackage{lineno,hyperref}
\usepackage{times}
\usepackage{epsfig}
\usepackage{graphicx}
\usepackage{amsmath}
\usepackage{amssymb}
\usepackage{bm}
\usepackage[dvipsnames]{xcolor}

\newcommand{\etal}{\textit{et al}.}
\DeclareMathOperator*{\argmax}{arg\,max}

\modulolinenumbers[5]

\begin{document}

\title{Hybrid eye center localization using cascaded regression and hand-crafted model fitting}

\author{Alex Levinshtein$^1$\\
	{\tt\small alex@modiface.com}
	\and
	Edmund Phung$^1$ \\
	{\tt\small edmund@modiface.com} \\
	\and
	Parham Aarabi$^{1,2}$ \\
	{\tt\small parham@modiface.com} \\
	\and
	$^1${\small ModiFace Inc., Toronto, Canada}
	\and
	$^2${\small Department of Electrical and Computer Engineering, University of Toronto, Toronto, Canada} 
}

\maketitle

\begin{abstract}
We propose a new cascaded regressor for eye center detection. Previous methods start from a face or an eye detector and use either advanced features or powerful regressors for eye center localization, but not both. Instead, we detect the eyes more accurately using an existing facial feature alignment method. We improve the robustness of localization by using both advanced features and powerful regression machinery. Unlike most other methods that do not refine the regression results, we make the localization more accurate by adding a robust circle fitting post-processing step. Finally, using a simple hand-crafted method for eye center localization, we show how to train the cascaded regressor without the need for manually annotated training data. We evaluate our new approach and show that it achieves state-of-the-art performance on the BioID, GI4E, and the TalkingFace datasets. At an average normalized error of $e < 0.05$, the regressor trained on manually annotated data yields an accuracy of $95.07\%$ (BioID), $99.27\%$ (GI4E), and $95.68\%$ (TalkingFace). The automatically trained regressor is nearly as good, yielding an accuracy of $93.9\%$ (BioID), $99.27\%$ (GI4E), and $95.46\%$ (TalkingFace).
\end{abstract}

\section{Introduction}
\label{sec:Introduction}

\begin{figure*}[t]
	\begin{center}
		\includegraphics[width=0.95\linewidth]{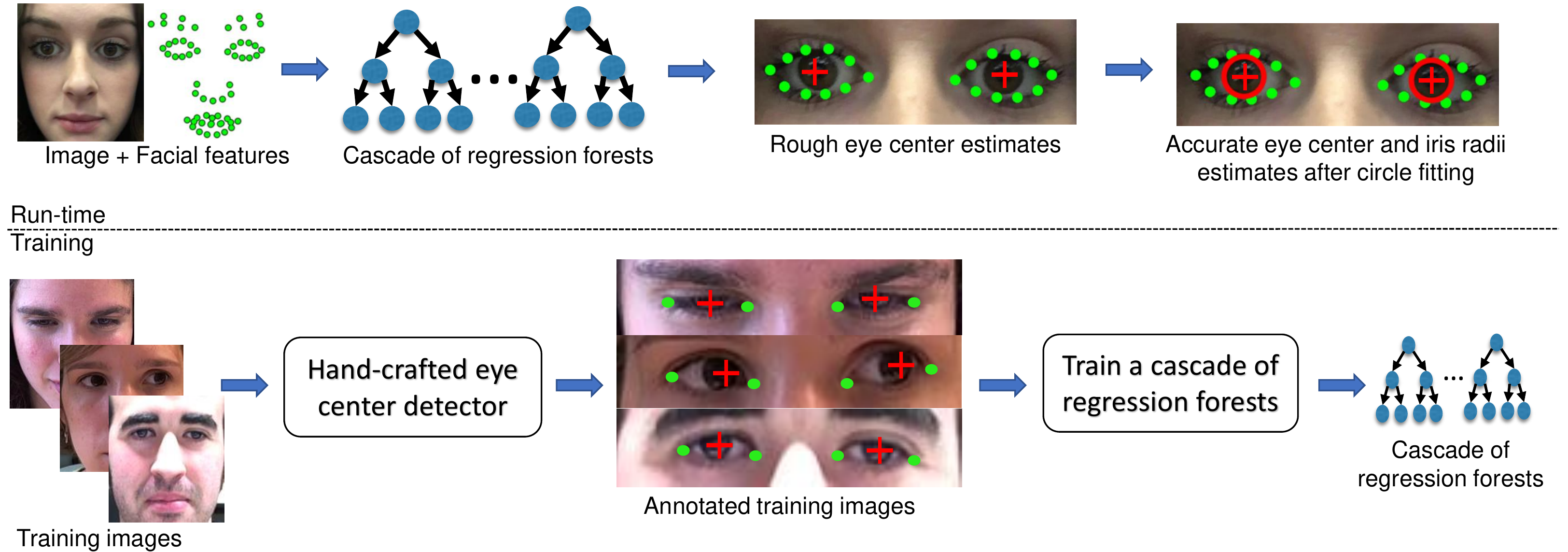}
	\end{center}
	\caption{Overview of our method.  Top: At run-time an image and the detected facial features are used to regress the eye centers, which are then refined by fitting circles to irises. Bottom: Rather than training the regressor from manually annotated data, we can train the regressor on automatically generated annotations from a hand-crafted eye center detector.}
	\label{fig:intro}
\end{figure*}

Eye center localization in the wild is important for a variety of applications, such as eye tracking, iris recognition, and more recently augmented reality applications for the beauty industry enabling virtual try out of contact lenses. While some techniques require the use of specialized head-mounts or active illumination, such machinery is expensive and is not applicable in many cases. In this paper, we focus on eye center localization using a standard camera. Approaches for eye center localization can be divided into two categories. The first, predominant, category consists of hand-crafted model fitting methods. These techniques employ the appearance, such as the darkness of the pupil, and/or the circular shape of the pupil and the iris for detection \cite{ahuja2016eye,fuhl2015excuse,fuhl2016else,george2016fast,li2005starburst,skodras2015precise,swirski2012robust,timm2011accurate,valenti2012accurate,wood2014eyetab}. These methods are typically accurate but often lack robustness in more challenging settings, such as low resolution or noisy images and poor illumination. More recently, a second category emerged - machine learning based methods. While there are approaches that train sliding window eye center detectors, recent success of cascaded regression for facial feature alignment \cite{cao2014face,ren2014face,kazemi2014one,xiong2013supervised} has prompted the community to apply these methods for eye center localization \cite{markuvs2014eye,tian2016accurate,zhou2015precise}. These new methods have proven to be more robust, but they lack the accuracy of the model fitting approaches and require annotated training data which may be cumbersome to obtain. 

In this paper, we propose a novel eye center detection method that combines the strengths of the aforementioned categories. In the literature on facial feature alignment, there are two types of cascaded regression methods, simple cascaded linear regressors using complex features such as SIFT or HoG \cite{antonakos2015feature,xiong2013supervised} and more complex cascaded regression forests using simple pairwise pixel difference features \cite{cao2014face,kazemi2014one}. Our first contribution is a new method for eye center localization that employs complex features \textbf{and} complex regressors. It outperforms simple regressors with complex features \cite{zhou2015precise} and  complex regressors with simple features \cite{markuvs2014eye,tian2016accurate}. Similar to \cite{markuvs2014eye,tian2016accurate} our method is based on cascaded regression trees, but unlike these authors, following \cite{cao2014face,ren2014face,kazemi2014one,xiong2013supervised}, our features for each cascade are anchored to the current eye center estimates. Moreover, based on the pedestrian detection work of Dollar \etal \cite{dollar2009integral} we employ more powerful gradient histogram features rather than simple pairwise pixel differences. Finally, while the aforementioned eye center regressors bootstrap the regression using face or eye detectors, given the success of facial feature alignment methods we make use of accurate eye contours to initialize the regressor and normalize feature locations. We show that the resulting method achieves state-of-the-art performance on BioID \cite{bioid}, GI4E \cite{gi4e}, and TalkingFace \cite{talkingface} datasets.

The proposed cascaded regression approach is robust, but suffers from the same disadvantages of other discriminative regression-based methods. Namely, it is inaccurate and requires annotated training data. To make our approach more accurate, in our second contribution we refine the regressor estimate by adding a circle fitting post-processing step. Employing robust estimation and prior knowledge of iris size facilitates sub-pixel accuracy eye center detection. We show the benefit of this refinement step by evaluating our approach on GI4E \cite{gi4e} and TalkingFace \cite{talkingface} datasets, as well as performing qualitative evaluation.

Finally, for our third contribution, rather than training our regressor on manually generated annotations we employ a hand-crafted method to generate annotated data automatically. Combining recent advances of eye center and iris detection methods, we build a new hand-crafted eye center localization method. It performs well compared to its hand-crafted peers, but is inferior to regressor-based approaches. Despite the noisy annotations generated by the hand-crafted algorithm, we show that the resulting regressor trained on these annotation is nearly as good as the regressor trained on manually annotated data. What is even more unexpected is that the regressor performs much better than the hand-crafted method used for training data annotation.

In summary, this paper proposes a new state-of-the-art method for eye center localization and has three main contributions that are shown in Figure \ref{fig:intro}. First, we present a novel cascaded regression framework for eye center localization, leading to increased robustness. Second, we add a circle fitting step, leading to eye center localization with sub-pixel accuracy. Finally, we show that by employing a hand-crafted eye center detector the regressor can be trained without the need for manually annotated training data. This paper builds on the work that was presented in a preliminary form in \cite{levinshteinSIP2017}.

\section{Related Work}
\label{sec:Related}

The majority of eye center localization methods are hand-crafted approaches and can be divided into shape and appearance based methods. In the iris recognition literature there are also many segmentation based approaches, such as methods that employ active contours. An extensive overview is given by Hansen and Li \cite{hansen2010eye}. Shape-based techniques make use of the circular or elliptical nature of the iris and pupil. Early methods attempted to detect irises or pupils directly by fitting circles or ellipses. Many techniques have roots in the iris recognition and are based on the integrodifferential operator \cite{daugman2004iris} . Others, such as Kawaguchi \etal \cite{kawaguchi2000detection}, use blob detection to extract iris candidates and use Hough transform to fit circles to these blobs. Toennies \etal \cite{toennies2002feasibility} also employ generalized Hough transform to detect irises, but assume that every pixel is a potential edge point and cast votes proportional to gradient strength. Li \etal \cite{li2005starburst} propose the Startburst algorithm, where rays are iteratively cast from the current pupil center estimate to detect pupil boundaries and RANSAC is used for robust ellipse fitting.

Recently, some authors focused on robust eye center localization without an explicit segmentation of the iris or the pupil. Typically, these are either voting or learning-based approaches. The method of Timm and Barth \cite{timm2011accurate} is a popular voting based approach where pixels cast votes for the eye center based on agreement in the direction of their gradient with the direction of radial rays. A similar voting scheme is suggested by Valenti and Gevers \cite{valenti2012accurate}, who also cast votes based on the aforementioned alignment but rely on isophote curvatures in the intensity image to cast votes at the right distance. Skodras and Fakotakis \cite{skodras2015precise} propose a similar method but use color to better distinguish between the eye and the skin. Ahuja \etal \cite{ahuja2016eye} improve the voting using radius constraints, better weights, and contrast normalization. 

The next set of methods are multistage approaches that first robustly detect the eye center and then refine the estimate using circle or ellipse fitting.
{\'S}wirski \etal \cite{swirski2012robust} propose to find the pupil using a cascade of weak classifiers based on Haar-like features combined with intensity-based segmentation. Subsequently, an ellipse is fit to the pupil using RANSAC. Wood and Bulling \cite{wood2014eyetab}, as well as George and Routray \cite{george2016fast},  have a similar scheme but employ a voting-based approach to get an initial eye center estimate. Fuhl \etal propose the Excuse \cite{fuhl2015excuse} and Else \cite{fuhl2016else} algorithms. Both methods use a combination of ellipse fitting with appearance-based blob detection.

While the above methods are accurate, they still lack robustness in challenging in-the-wild scenarios. The success of discriminative cascaded regression for facial feature alignment prompted the use of such methods for eye center localization. \cite{markuvs2014eye,tian2016accurate} start by detecting the face and initializing the eye center estimates using anthropometric relations. Subsequently, they use a cascade of regression forests with binary pixel difference features to estimate the eye centers. Inspired by the recent success of the SDM method for facial feature alignment Zhou \etal \cite{zhou2015precise} propose a similar method for eye center localization. Unlike the original SDM work, their regressor is based on a combination of SIFT and LBP features. Moreover, unlike \cite{markuvs2014eye,tian2016accurate} who regress individual eye centers, Zhou \etal estimate a shape vector that includes both eye centers and eye contours. In line with this trend we develop a new regression-based eye center estimator, but additionally employ circle-based refinement and voting-based techniques to get an accurate detector that is easy to train.

\section{Eye Center Localization}
\label{sec:Algorithm}

In this section, we describe our three main contributions in detail. We start by introducing our cascaded regression framework for eye center localization (Section \ref{subsec:Regressor}). Next, we show how the eye center estimate can be refined with a robust circle fitting step by fitting a circle to the iris (Section \ref{subsec:Circlefitting}). Section \ref{subsec:Autotraining} explains how to train the regressor without manually annotated eye center data by using a hand-crafted method for automatic annotation. Finally, in Section \ref{subsec:Closedeyes} we discuss our handling of closed or nearly closed eyes.

\subsection{Cascaded regression framework}
\label{subsec:Regressor}
Inspired by the face alignment work in \cite{cao2014face,kazemi2014one}, we build an eye center detector using a cascade of regression forests. Our shape is represented by a vector $\bm{S} = (\bm{x}_R^T, \bm{x}_L^T)$, where $\bm{x}_R$ and $\bm{x}_L$ are the coordinates of right and left eye centers respectively. Starting from an initial guess $\bm{S}^0$, we refine the shape using a cascade of regression forests:
\begin{equation}
\bm{S}^{t+1} = \bm{S}^{t} + r_t(I, \bm{S}^t), 
\end{equation}
\noindent
where $r_t$ is the $t$-th regressor in the cascade estimating the shape update given the image $I$ and the current shape estimate $\bm{S}^t$. Next, we describe our choice of image features, our regression machinery, and the mechanism for obtaining an initial shape estimate $\bm{S}^0$.


For our choice of image features, similar to Dollar \etal \cite{dollar2009integral}, we selected HoG features anchored to the current shape estimate. We find that using HoG is especially helpful for bright eyes, where variation in appearance due to different lighting and image noise is more apparent, hurting the performance of regressors employing simple pixel difference features. Zhou \etal \cite{zhou2015precise}, also employ advanced image features, but in contrast to them we use regression forests at each level of our cascade. Finally, while \cite{markuvs2014eye,tian2016accurate} estimate eye center positions independently, we find that due to the large amount of correlation between the two eyes it is beneficial to estimate both eyes jointly. In \cite{zhou2015precise}, the shape vector consists of eye centers and their contours. However, since it is possible to change gaze direction without a change in eye contours, our shape vector $S$ includes only the two eye center points.

\begin{figure}[t]
	\begin{center}
		\includegraphics[width=0.95\linewidth]{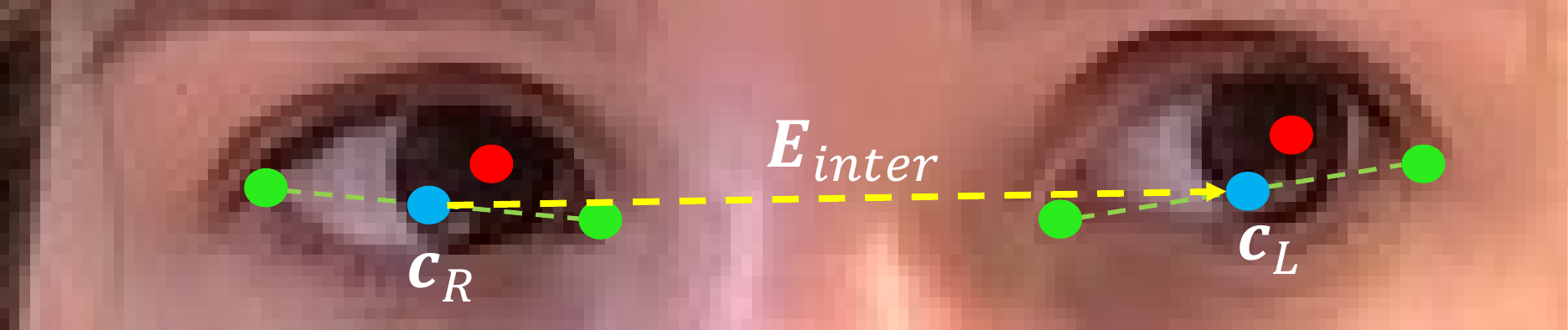}
	\end{center}
	\caption{Normalized eye center locations. The eye centers (red) coordinates are normalized such that the vector $\bm{E}_{inter}$ (yellow) between the two eye centers $\bm{c}_R$ and $\bm{c}_L$ (cyan) maps to the vector $(1,0)^T$.}
	\label{fig:normshape}
\end{figure}

To get an initial shape estimate, existing approaches use eye detectors or face detectors with anthropometric relations to extract the eye regions. Instead, we employ a facial feature alignment method to get an initial shape estimate and anchor features. Specifically, the four eye corners are used to construct a normalized representation of shape $S$. We define $\bm{c}_R$ and $\bm{c}_L$, to be the center points between the corners of the right and left eyes respectively. The vector $\bm{E}_{inter}$ between the two eye centers is defined as the interocular vector with its magnitude $\lVert \bm{E}_{inter} \rVert$ defined as the interocular distance. Figure \ref{fig:normshape} illustrates this geometry. The similarity transformation $T(\bm{x})$ maps points from image to face-normalized coordinates and is defined to be the transformation mapping $\bm{E}_{inter}$ to a unit vector aligned with the $X$ axis with the $\bm{c}_R$ mapped to the origin. Therefore, the shape vector $S$ consists of two normalized eye centers $\bm{x}_R = T(\bm{x}_R^{image})$ and $\bm{x}_L = T(\bm{x}_L^{image})$ with $\bm{x}_R^{image}$ and $\bm{x}_L^{image}$ being the eye center estimates in the image. The eye center estimates $\bm{c}_R$ and $\bm{c}_L$ are also used to define the initial shape $\bm{S^0} = (T(\bm{c}_R)^T, T(\bm{c}_L)^T)$.

At each level of the cascade, we extract HoG features centered at the current eye center estimates.
To make HoG feature extraction independent of the face size we scale the image by a factor $s = \frac{E_{hog}}{\lVert \bm{E}_{inter} \rVert}$, where $E_{hog}$ is the constant interocular distance used for HoG computation. Using bilinear interpolation, we extract $W \times W$ patches centered at the current eye center estimates $sT^{-1}(\bm{x}_R)$ and $sT^{-1}(\bm{x}_L)$, with $W = 0.4 E_{hog}$. Both patches are split into $4 \times 4$ HoG cells with 6 oriented gradient histogram bins per cell. The cell histograms are concatenated and the resulting vector normalized to a unit $L2$ norm, yielding a $96$ dimensional feature vector for each eye. Instead of using these features directly at the decision nodes of regression trees, we use binary HoG difference features. Specifically, at each decision node we generate a pool of $K$ ($K$ = 20 in our implementation) pairwise HoG features by randomly choosing an eye, two of the $96$ HoG dimensions, and a threshold. The binary HoG difference feature is defined as the thresholded difference between the chosen HoG components. During training, the feature that minimizes the regression error is selected.

To train the cascaded regressor, we use a dataset of annotated images with eye corners and centers. To model the variability in eye center locations we use Principal Components Analysis. Using a simple form of Procrustes Analysis, each training shape is translated to the mean shape and the resulting shapes are used to build a PCA basis. Subsequently, for each training image, multiple initial shapes $S^0$ are sampled by generating random PCA coefficients, centering the resulting shape at the mean, and translating both eyes by the same random amount. The random translation vector is sampled uniformly from the range $\big[ -0.1,0.1 \big]$ in $X$ and $\big[ -0.03,0.03 \big]$ in $Y$. The remaining parameters of the regressor are selected using cross validation. Currently, our regressor has $10$ levels with $200$ depth-$4$ trees per level. Each training image is oversampled $50$ times. The regressor is trained using gradient boosting, similar to \cite{kazemi2014one}, with the learning rate parameter set to $\nu = 0.1$.

\subsection{Iris refinement by robust circle fitting}
\label{subsec:Circlefitting}

We refine the eye center position from the regressor by fitting a circle to the iris. Our initial circle center is taken from the regressor and the radius estimate $r_{init}$ starts with a default of $0.1 \lVert \bm{E}_{inter} \rVert$. The iris is refined by fitting a circle to the iris boundaries. To that end, assuming the initial circle estimate is good enough, we extract edge points that are close to the initial circle boundary as candidates for the iris boundary. Not all of the extracted edge points will belong to the iris boundary and our circle fitting method will need to handle these outliers.

\begin{figure}[t]
	\begin{center}
		\includegraphics[width=0.95\linewidth]{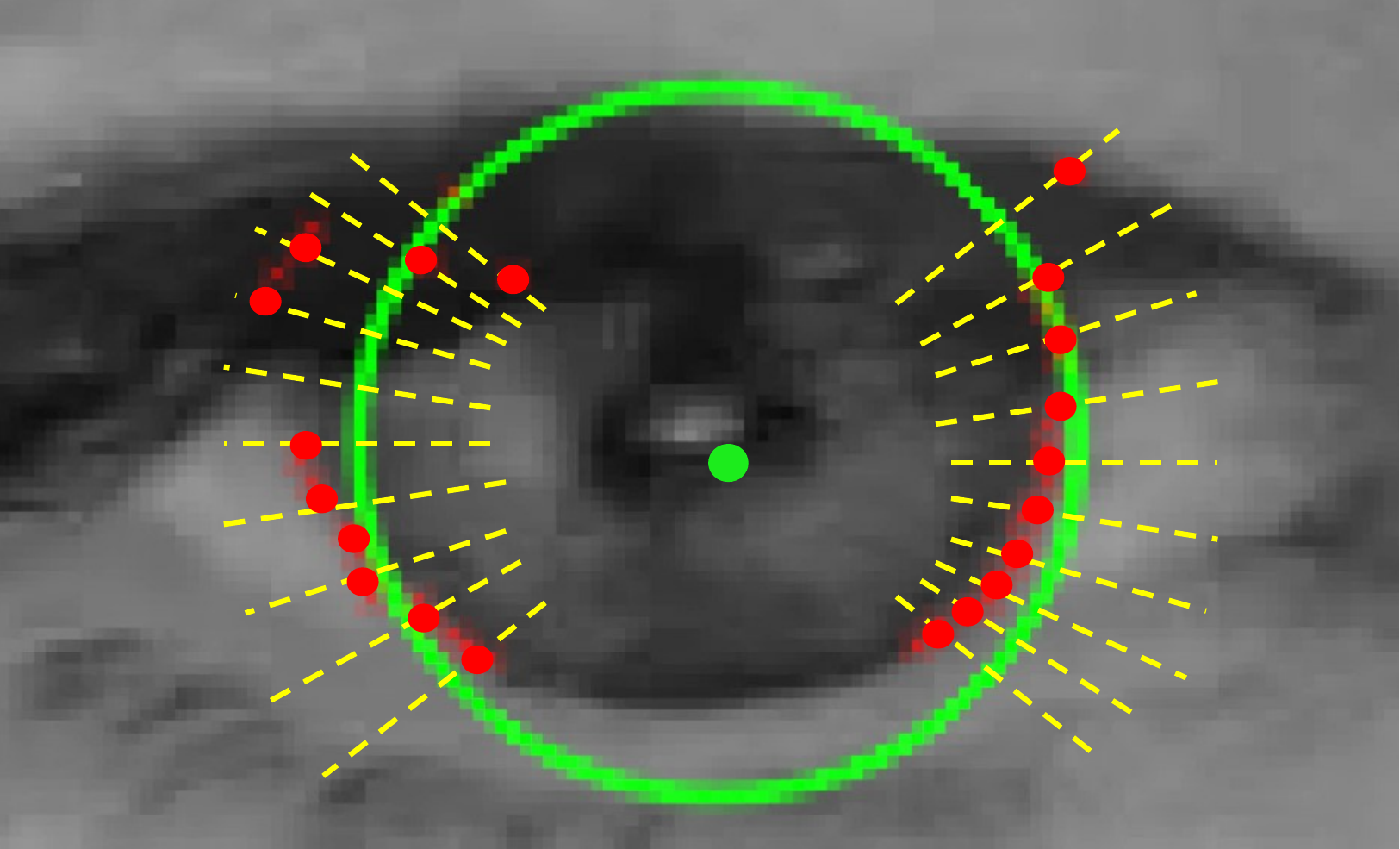}
	\end{center}
	\caption{Robust circle fitting for iris refinement. Starting from an initial iris estimate (green center and circle), short scan lines (yellow) perpendicular to the initial circle are used to detect candidate iris boundaries (red). A robust circle fitting method is then used to refine the estimate.}
	\label{fig:circlefitting}
\end{figure}

Employing the eye contours once again, we start by sampling $N$ points on the circle and removing the points that lie outside the eye mask. To avoid extracting the eyelid edges we only consider circle samples in range $\pm 45^{\circ}$ and $\big[ 135^{\circ},225^{\circ} \big]$. For each circle point sample we form a scan line centered on that point and directed toward the center of the circle. The scan line is kept short ($\pm 30\%$ of the circle radius) to avoid extracting spurious edges. Each point on the scan line is assigned a score equal to the dot product between the gradient and outwards-facing circle normal. The highest scoring point location is stored. Points for which the angle between the gradient and the normal is above $25^{\circ}$ are not being considered. This process results in a list of edge points (red points in Figure \ref{fig:circlefitting}).  

Given the above edge points $\big\{ \bm{e}_i \big\}_{i=1}^{N}$, the circle fitting cost is defined as follows:
\begin{equation}
\label{eqn:circle_not_robust}
C(a,b,r) = \sum_{i=1}^{N} \Big( \sqrt{(e_{ix} - a)^2 + (e_{iy} - b)^2} - r \Big)^2 ,
\end{equation}

\noindent
where $(a,b)$ is the circle center and $r$ is the radius. However, this cost is not robust to outliers nor are any priors for circle location and size being considered. Thus, we modify the cost to the following:

\begin{align}
\label{eqn:circle_robust}
C_2 &= w_1 \cdot \frac{1}{N} \sum_{i=1}^{N} \rho \Big( \sqrt{(e_{ix} - a)^2 + (e_{iy} - b)^2} - r \Big) \notag \\ 
&+ w_2 \cdot \big(a - a_0)^2 \notag \\ 
&+ w_2 \cdot \big(b - b_0)^2 \notag \\ 
&+ w_3 \cdot \big(r - r_{default})^2 . 
\end{align}

Note that the squared cost in the first term was converted to a robust cost (we chose $\rho$ to be the Tukey robust estimator function). The rest are prior terms, where $(a_0,b_0)$ is the center estimate from the regressor and $r_{default} = 0.1 \lVert \bm{E}_{inter} \rVert$. We set the weights to $w_1 = 1, w_2 = 0.1, w_3 = 0.1$ and minimize the cost using the Gauss-Newton method with iteratively re-weighted least squares. The minimization process terminates if the relative change in cost is small enough or if a preset number of iterations (currently $30$) was exceeded. For the Tukey estimator, we start by setting its parameter $C = 0.3 r_{init}$ and decrease it to $C = 0.1 r_{init}$ after initial convergence. 

\subsection{Using a hand-crafted detector for automatic annotations}
\label{subsec:Autotraining}
As mentioned in Section \ref{sec:Related}, there are a variety of hand-crafted techniques for eye center localization. Some methods work well in simple scenarios but are still falling short in more challenging cases. In this section, we construct our own hand-crafted method for eye center localization and use it to automatically generate annotations for a set of training images. The resulting annotations can be considered as noisy training data. One can imagine similar data as the output of a careless human annotator. We then train the cascaded regressor from Section \ref{subsec:Regressor} on this data. Since the output of the regressor is a weighted average of many training samples, it naturally smooths the noise in the annotations and yields better eye center estimates than the hand-crafted method used to generate the annotations. Next, we describe the approach in more detail.

Our hand-crafted eye center localization method is based on the work of Timm and Barth \cite{timm2011accurate}. Since we are looking for circular structures, \cite{timm2011accurate} propose finding the maximum of an eye center score function $S(\bm{c})$ that measures the agreement between vectors from a candidate center point $\bm{c}$ and underlying gradient orientation:
\begin{equation}
\label{eqn:TimmScore}
\bm{c^*} = \argmax_{\bm{c}} S(\bm{c}) = \argmax_{\bm{c}} { \Bigg\{ \frac{1}{N}\sum_{i=1}^{N} {w_{\bm{c}} \left( \bm{d}_i^T \bm{g}_i \right) ^2} \Bigg\} },
\end{equation}

\noindent
where $\bm{d}_i$ is the normalized vector from $\bm{c}$ to point $i$ and $\bm{g}_i$ is the normalized image gradient at $i$. $w_{\bm{c}}$ is the weight of a candidate center $\bm{c}$. Since the pupil is dark, $w_{\bm{c}}$ is high for dark pixels and low otherwise. Specifically, $w_{\bm{c}} = 255 - I^*(\bm{c})$, where $I^*$ is an 8-bit smoothed grayscale image. 

Similar to \cite{ahuja2016eye}, we observe that an iris has a constrained size. More specifically, we find that its radius is about $20\%$ of the eye size $E$, which we define as the distance between the two eye corners. Thus, we only consider pixels $i$ within a certain range of $c$. Furthermore, the iris is darker than the surrounding sclera. The resulting score function is:

\begin{equation}
\label{eqn:OurScore}
S(\bm{c}) = \frac{1}{N}\sum_{0.3 E \le \lVert \bm{d}_i^* \rVert \le 0.5 E} {w_{\bm{c}} \max \left( \bm{d}_i^T \bm{g}_i , 0 \right)},
\end{equation}

\noindent
where $\bm{d}_i^*$ is the unnormalized vector from $\bm{c}$ to $i$. 

Unlike \cite{timm2011accurate} that find the global maximum of the score function in Eqn \ref{eqn:TimmScore}, we consider several local maxima of our score function as candidates for eye center locations. To constrain the search, we use a facial feature alignment method to obtain an accurate eye mask. We erode this mask to avoid the effect of eye lashes and eyelids, and find all local maxima of $S(\bm{c})$ in Eqn \ref{eqn:OurScore} within the eroded eye mask region whose value is above $80\%$ of the global maximum. Next, we refine each candidate and select the best one.

Since each candidate's score has been accumulated over a range of radii, starting with a default iris radius of $0.2E$, the position and the radius of each candidate is refined. The refinement process evaluates the score function in an 8-connected neighborhood around the current estimate. However, instead of summing over a range of radii as in Eqn \ref{eqn:OurScore}, we search for a single radius that maximizes the score. Out of all the 8-connected neighbors together with the central position, we select the location with the maximum score and update the estimate. The process stops when all the 8-connected neighbors have a lower score than the central position. Finally, after processing all eye center candidates in the above fashion, we select a single highest scoring candidate. Its location and radius estimates are then refined to sub-pixel accuracy using the robust circle fitting method from Section \ref{subsec:Circlefitting}.

In the next step, we use our hand-crafted method to automatically annotate training images. Given a set of images, we run the facial feature alignment method and the hand-crafted eye center detector on each image. We annotate each image with the position of the four eye corners from the facial feature alignment method and the two iris centers from our hand-crafted detector. Finally, we train the regressor from Section \ref{subsec:Regressor} on this data. In Section \ref{sec:Evaluation} we show that the resulting regressor performs much better than the hand-crafted method on both training and test data, and performs nearly as well as a regressor trained on manually annotated images.

\subsection{Handling closed eyes}
\label{subsec:Closedeyes}

Our algorithm has the benefit of having direct access to eye contours for estimating the amount of eye closure. To that end, we fit an ellipse to each eye's contour and use its height to width ratio $r$ to control our algorithm flow. For $r > 0.3$, which holds for the majority of cases we have examined, we apply both the regression and the circle fitting methods described in previous sections. For reliable circle refinement, a large enough portion of the iris boundary needs to be visible. Thus, for $0.15 < r \le 0.3$ we only use the regressor's output. For $r \le 0.15$ we find even the regressor to be unreliable, thus the eye center is computed by averaging the central contour points on the upper and lower eyelids.

\section{Evaluation}
\label{sec:Evaluation}
We perform quantitative and qualitative evaluation of our method and compare it to other approaches. For quantitative evaluation we use the normalized error measure defined as:

\begin{equation}
\label{eqn:error}
e = \frac{1}{d}\max(e_R, e_L),
\end{equation}

\noindent
where $e_R$, $e_L$ are the Euclidean distances between the estimated and the correct right and left eye centers, and $d$ is the distance between the correct eye centers. When analyzing the performance, different thresholds on $e$ are used to assess the level of accuracy. The most popular metric is the fraction of images for which $e \le 0.05$, which roughly means that the eye center was estimated somewhere within the pupil. In our analysis we pay closer attention to even finer levels of accuracy as they may be needed for some applications, such as augmented beauty or iris recognition, where the pupil/iris need to be detected very accurately.

While there are many public facial datasets available, many do not contain iris labels. Others may have iris only images or NIR images. For our analysis, we require color or grayscale images containing the entire face with labeled iris centers. Therefore, we use the BioID \cite{bioid}, GI4E \cite{gi4e}, and TalkingFace \cite{talkingface} datasets for evaluation. The BioID dataset consists of $1521$ low resolution ($384 \times 286$) images. Images exhibit wide variability in illumination and contain several closed or nearly closed eyes. While this dataset tests the robustness of eye center detection, its low resolution and the presence of closed eyes make it less suitable to test the fine level accuracy (finer levels than $e \le 0.05$). The GI4E and the Talking Face datasets have $1236$ and $5000$ high resolution images respectively and contain very few closed eye images. Thus, we find these datasets to be more appropriate for fine level accuracy evaluation.

We implement our method in C/C++ using OpenCV and DLIB libraries. Our code takes $4$ms to detect both eye centers on images from the BioID dataset using a modern laptop computer with Xeon $2.8$GHz CPU, not including the face detection and face alignment time. The majority of this time is spent on image resizing and HoG feature computation using the unoptimized code in the DLIB library and can be significantly sped up. Given the features, traversing the cascade is fast. For each $4$-level tree, $3$ subtractions and comparisons are needed to reach the leaf. The shifts ($4$ values for the two iris center coordinates) in all the trees are added together to compute the final regressor output. This results in $3 \times 200 \times 10 = 6000$ subtractions and comparisons, and $4 \times 200 \times 10 = 8000$ floating point additions per image. The facial alignment method we use is based on \cite{kazemi2014one} and is part of the DLIB library, but any approach could be used for this purpose. Similar to previous methods, which rely on accurate face detection for eye center estimation, we require accurate eye contours for this purpose. To that end, we implemented a simple SVM-based approach for verifying alignment. Similar to previous methods, which evaluate eye center localization only on images with detected faces, we evaluate our method only on images for which the alignment was successful. While the alignment is successful in the vast majority of cases, some detected faces do not have an accurate alignment result. After filtering out images without successful alignment we are left with $1459/1521$ images ($95.9\%$) of the BioID dataset, $1235/1236$ images of the GI4E dataset, and all the $5000$ frames in the Talking Face dataset.

\subsection{Quantitative Evaluation}

\begin{figure*}[t!]
	\begin{minipage}{0.5\linewidth}
		\begin{center}
		\includegraphics[width=0.95\linewidth]{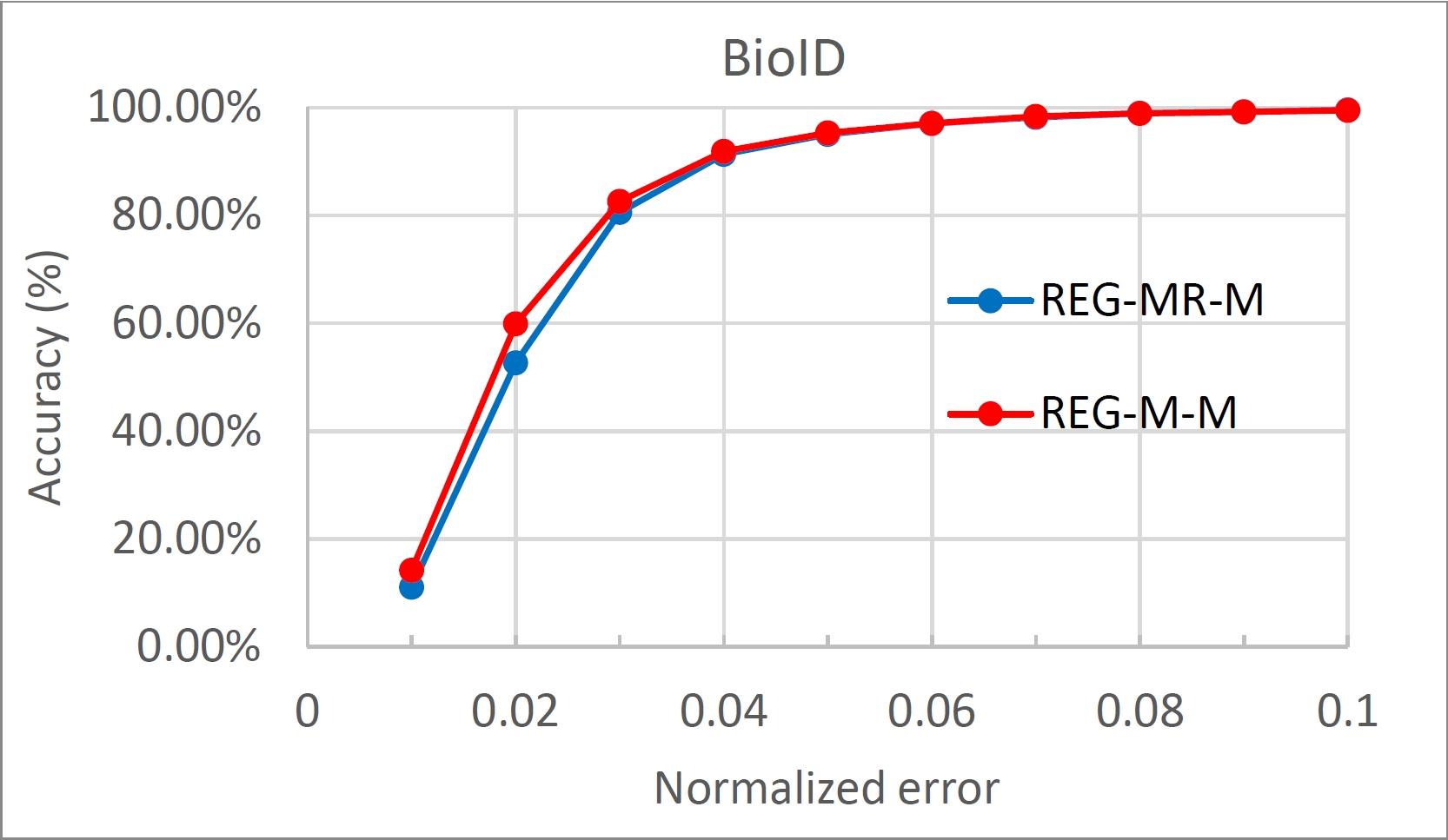} \\
		\includegraphics[width=0.95\linewidth]{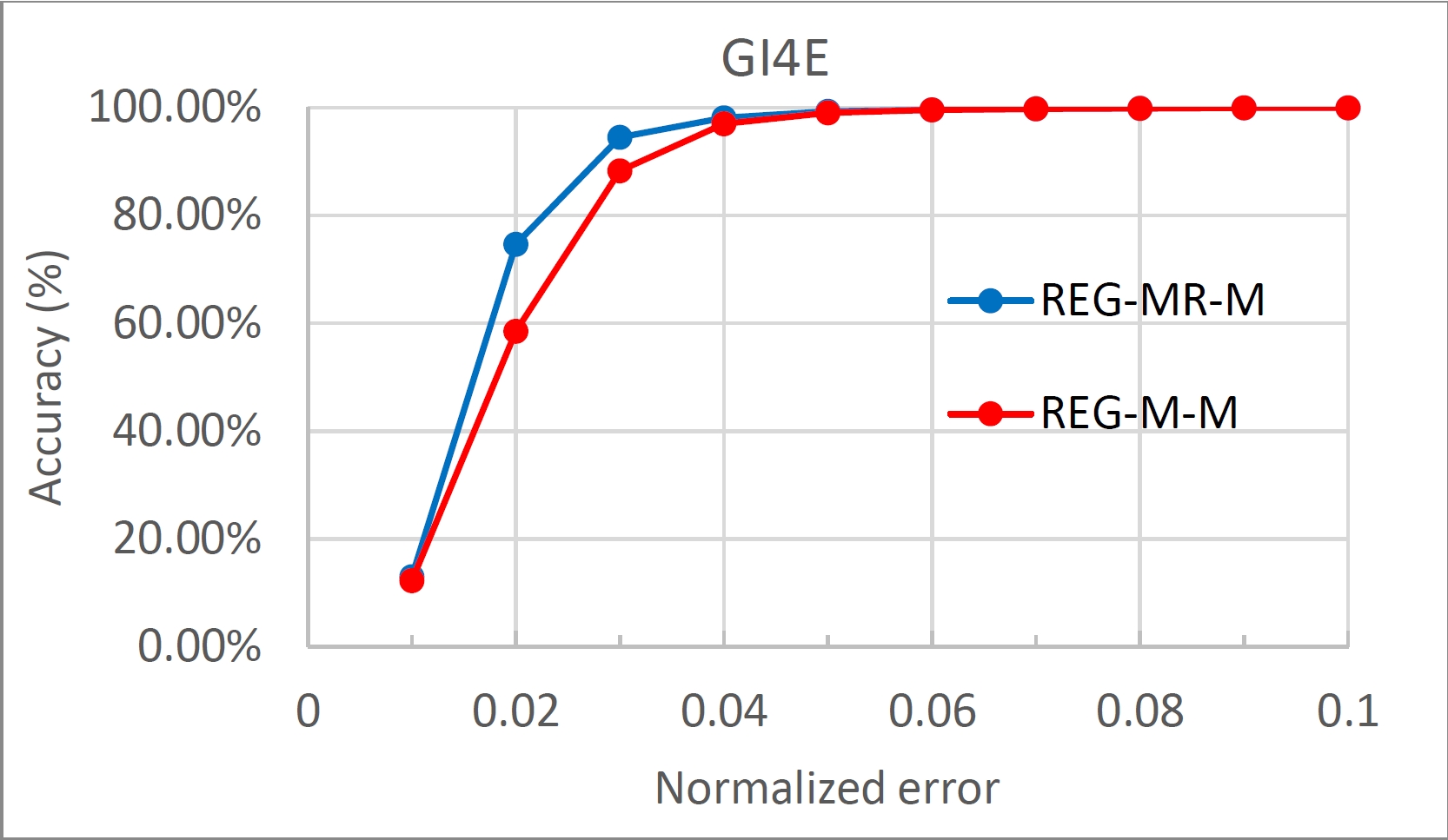} \\
		\includegraphics[width=0.95\linewidth]{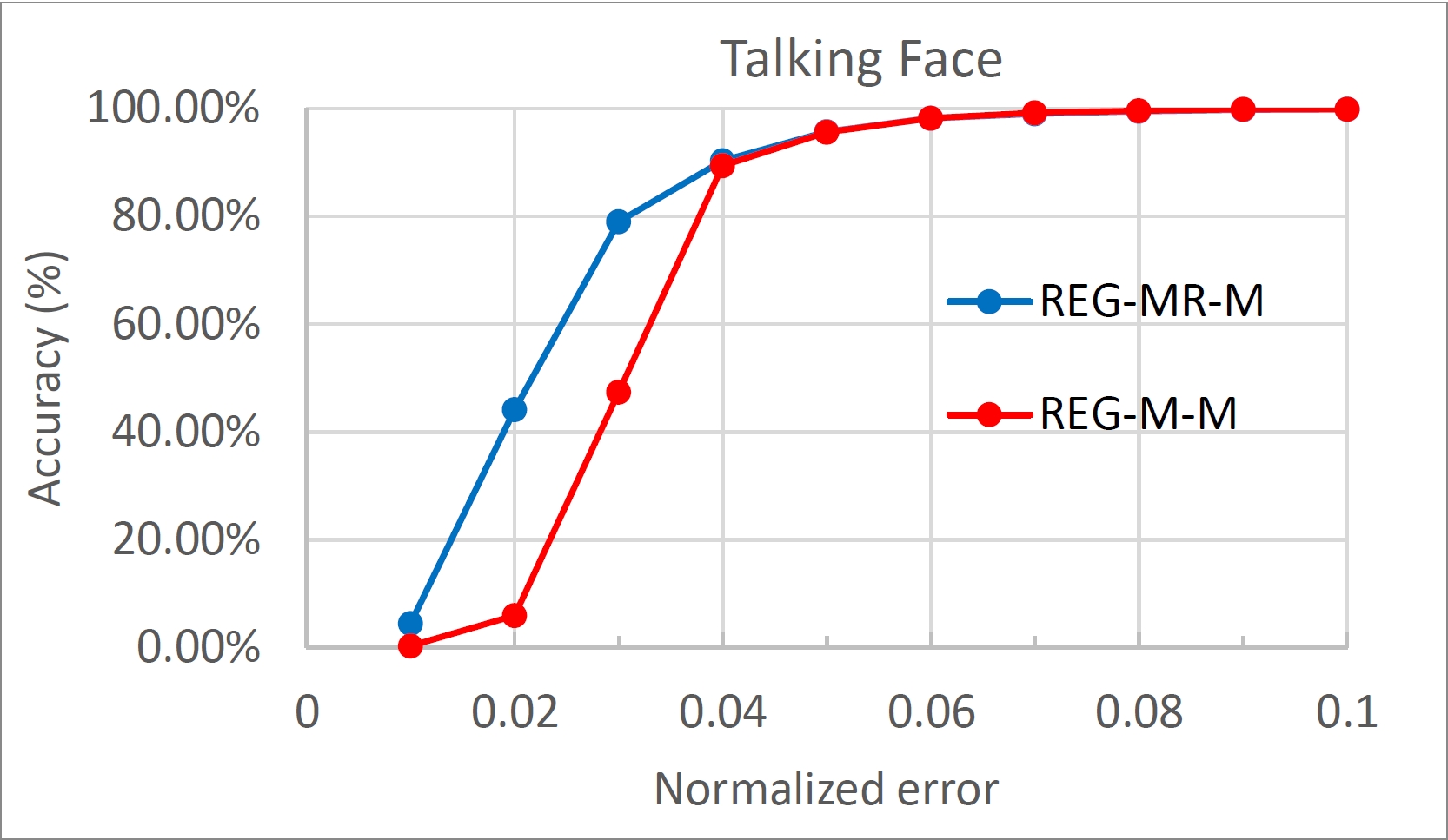} \\
		\mbox{\footnotesize Refined/Unrefined comparison}
		\end{center}
	\end{minipage}
	\begin{minipage}{0.5\linewidth}
		\begin{center}
		\includegraphics[width=0.95\linewidth]{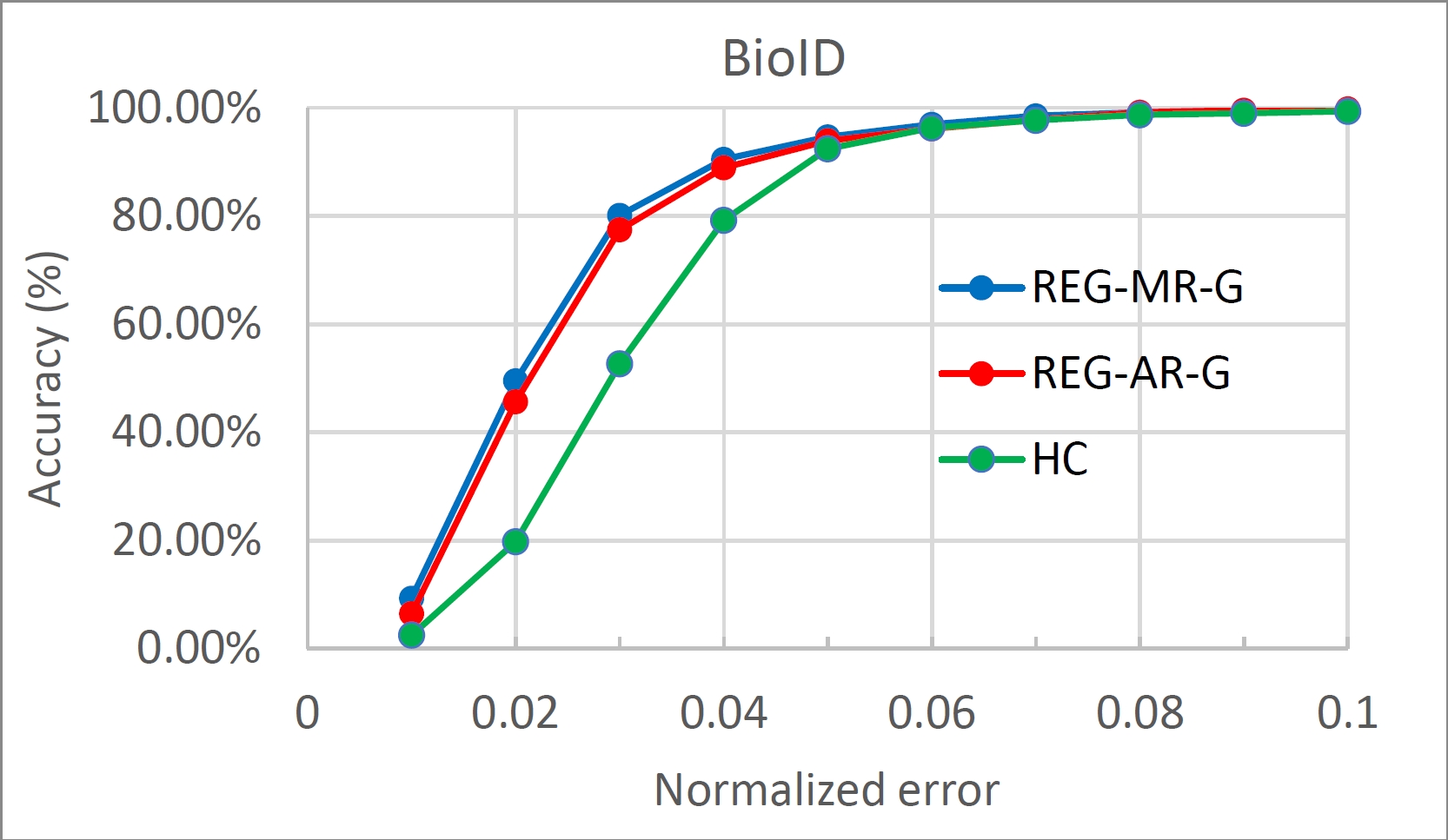} \\
		\includegraphics[width=0.95\linewidth]{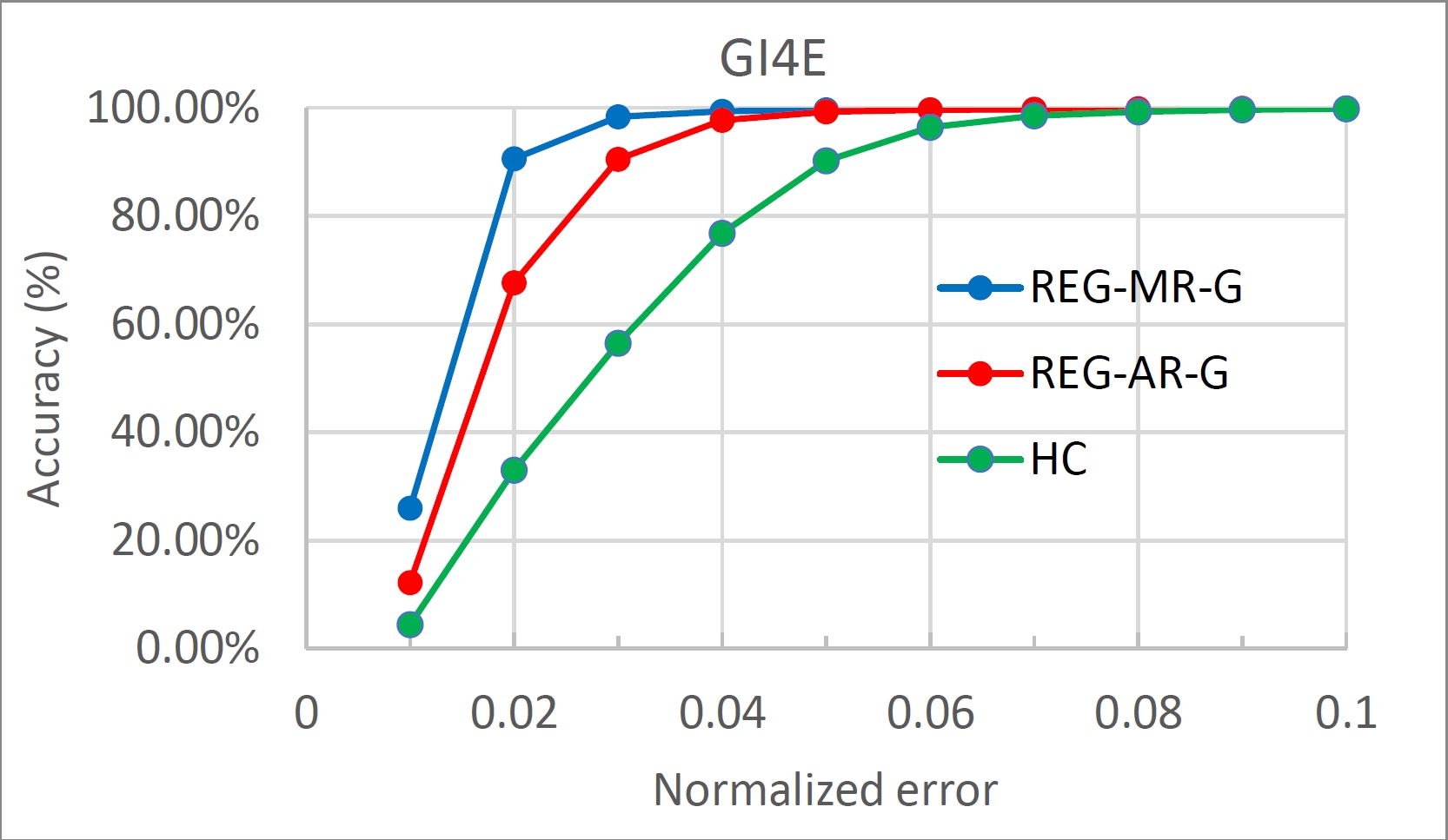} \\
		\includegraphics[width=0.95\linewidth]{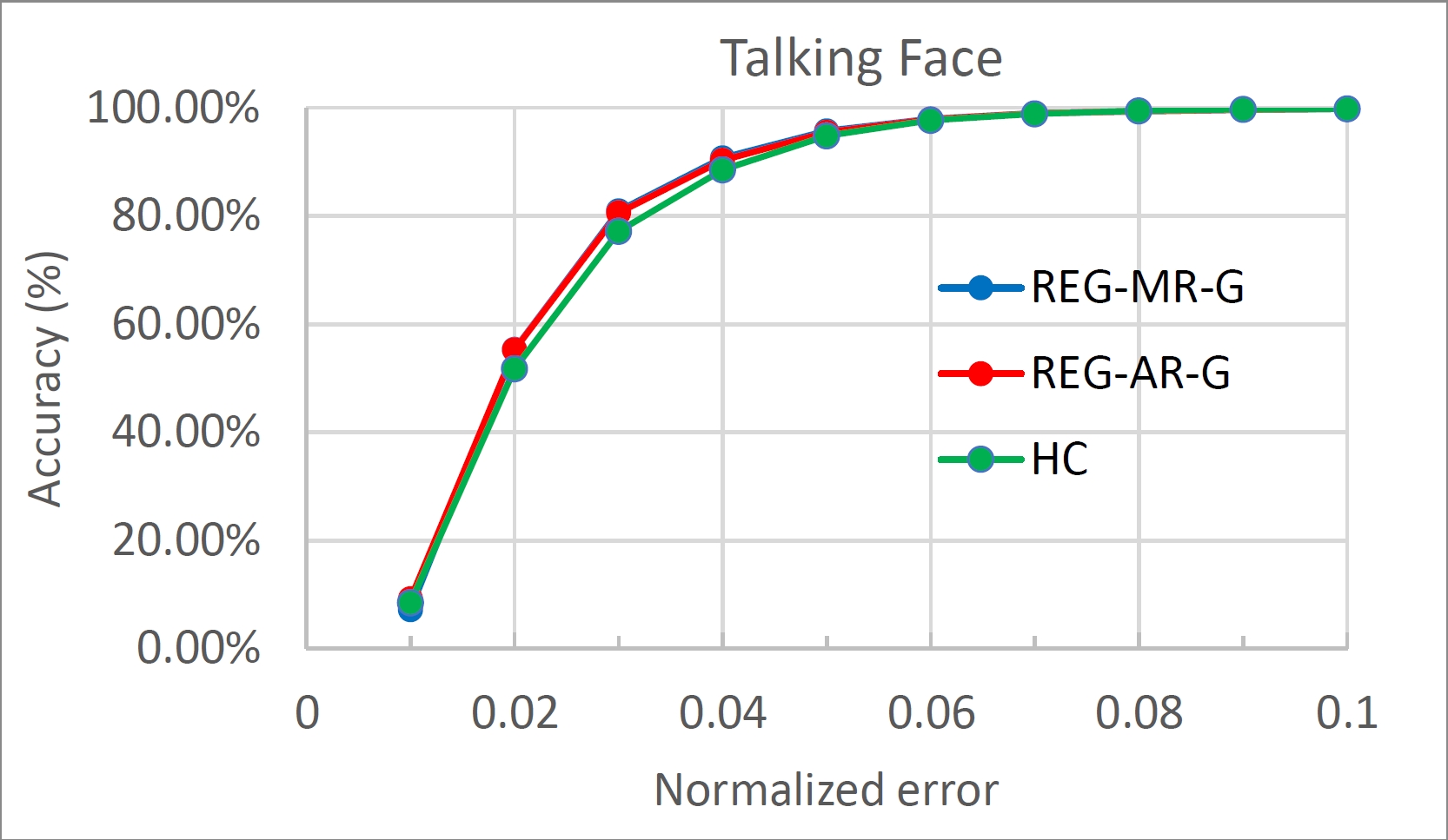} \\
		\mbox{\footnotesize Manual/Auto/HC comparison}
		\end{center}
	\end{minipage}
	\caption{Quantitative evaluation. Left: Evaluation of regression-based eye center detector trained on manually annotated MPIIGaze data with ({\textbf{\footnotesize REG-MR-M}}) and without ({\textbf{\footnotesize REG-M-M}}) circle refinement. Right: Comparison of automatically trained regressor ({\textbf{\footnotesize REG-AR-G}}) to manually trained regressor ({\textbf{\footnotesize REG-MR-G}}) and the hand-crafted method ({\textbf{\footnotesize HC}}) used to generate automatic annotations. }
	\label{fig:quantitative_evaluation_reg_man}
\end{figure*}

\begin{table*}[t!]
	\centering
	\setlength\tabcolsep{4pt}
	\begin{tabular}{|c||c|c|c|c|}
		\hline 
		\textbf{Method} & $\bm{e < 0.025}$ & $\bm{e < 0.05}$ & $\bm{e < 0.1}$ & $\bm{e < 0.25}$ \\ 
		\hline
		\multicolumn{5}{|c|}{BioID Dataset} \\
		\hline
		{\textbf{\footnotesize REG-MR-M}} & \textcolor{red}{$68.13\%$} & \textcolor{Orange}{$95.07\%$} & $99.59\%$ & \textcolor{ForestGreen}{$100\%$} \\ 
		\hline
		{\textbf{\footnotesize REG-M-M}} & \textcolor{ForestGreen}{$74.3\%$} & \textcolor{ForestGreen}{$95.27\%$} & $99.52\%$ & \textcolor{ForestGreen}{$100\%$} \\ 
		\hline
		{\textbf{\footnotesize REG-MR-G}} & \textcolor{Orange}{$68.75\%$} & \textcolor{red}{$94.65\%$} & \textcolor{red}{$99.73\%$} & \textcolor{ForestGreen}{$100\%$} \\ 
		\hline
		{\textbf{\footnotesize REG-AR-G}} & $64.15\%$ & $93.9\%$ & \textcolor{Orange}{$99.79\%$} & \textcolor{ForestGreen}{$100\%$} \\ 
		\hline
		{\textbf{\footnotesize HC}} & $36.26\%$ & $92.39\%$ & $99.38\%$ & \textcolor{ForestGreen}{$100\%$} \\
		\hline
		Timm and Barth \cite{timm2011accurate} & $38\%^*$ & $82.5\%$ & $93.4\%$ & $98\%$ \\
		\hline
		Valenti and Gevers \cite{valenti2012accurate} & $55\%^*$ & $86.1\%$ & $91.7\%$ & $97.9\%$ \\
		\hline
		Zhou et al. \cite{zhou2015precise} & $50\%^*$ & $93.8\%$ &\textcolor{ForestGreen}{$99.8\%$} & \textcolor{Orange}{$99.9\%$} \\
		\hline
		Ahuja et al. \cite{ahuja2016eye} & $NA$ & $92.06\%$ & $97.96\%$ & \textcolor{ForestGreen}{$100\%$} \\
		\hline
		Marku{\v{s}} et al. \cite{markuvs2014eye} & $61\%^*$ & $89.9\%$ & $97.1\%$ & \textcolor{red}{$99.7\%$} \\
		\hline
		\multicolumn{5}{|c|}{GI4E Dataset} \\
		\hline
		{\textbf{\footnotesize REG-MR-M}} & \textcolor{ForestGreen}{$88.34\%$} & \textcolor{ForestGreen}{$99.27\%$} & \textcolor{ForestGreen}{$99.92\%$} & \textcolor{ForestGreen}{$100\%$} \\ 
		\hline
		{\textbf{\footnotesize REG-M-M}} & \textcolor{red}{$77.57\%$} & \textcolor{Orange}{$99.03\%$} & \textcolor{ForestGreen}{$99.92\%$} & \textcolor{ForestGreen}{$100\%$} \\ 
		\hline
		{\textbf{\footnotesize REG-AR-G}} & \textcolor{Orange}{$83.32\%$} &\textcolor{ForestGreen}{$99.27\%$} & \textcolor{ForestGreen}{$99.92\%$} & \textcolor{ForestGreen}{$100\%$} \\ 
		\hline
		{\textbf{\footnotesize HC}} & $47.21\%$ & $90.2\%$ & \textcolor{Orange}{$99.84\%$} & \textcolor{ForestGreen}{$100\%$} \\
		\hline
		Fuhl et al. \cite{fuhl2016else} & $49.8\%$ & \textcolor{red}{$91.5\%$} & \textcolor{red}{$97.17\%$} & \textcolor{Orange}{$99.51\%$} \\
		\hline
		George and Routray \cite{george2016fast} & $NA$ & $89.28\%$ & $92.3\%$ & $NA$ \\	
		\hline
		\multicolumn{5}{|c|}{Talking Face Dataset} \\
		\hline
		{\textbf{\footnotesize REG-MR-M}} & $65.78\%$ & \textcolor{Orange}{$95.68\%$} & \textcolor{ForestGreen}{$99.88\%$} & \textcolor{ForestGreen}{$99.98\%$} \\ 
		\hline
		{\textbf{\footnotesize REG-M-M}} & $18.7\%$ & \textcolor{red}{$95.62\%$} & \textcolor{ForestGreen}{$99.88\%$} & \textcolor{ForestGreen}{$99.98\%$} \\ 
		\hline
		{\textbf{\footnotesize REG-MR-G}} & \textcolor{ForestGreen}{$71.56\%$} & \textcolor{ForestGreen}{$95.76\%$} & \textcolor{Orange}{$99.86\%$} & \textcolor{ForestGreen}{$99.98\%$} \\ 
		\hline
		{\textbf{\footnotesize REG-AR-G}} & \textcolor{Orange}{$71.16\%$} & $95.46\%$ & $99.82\%$ & \textcolor{ForestGreen}{$99.98\%$} \\ 
		\hline
		{\textbf{\footnotesize HC}} & \textcolor{red}{$67.74\%$} & $94.86\%$ & \textcolor{red}{$99.84\%$} & \textcolor{ForestGreen}{$99.98\%$} \\
		\hline
		ELSE \cite{fuhl2016else} & $59.26\%$ & $92\%$ & $98.98\%$ & \textcolor{Orange}{$99.94\%$} \\
		\hline
		Ahuja \cite{ahuja2016eye} & $NA$ & $94.78\%$ & $99\%$ & \textcolor{red}{$99.42\%$} \\
		\hline
	\end{tabular} 
	\caption{Quantitative evaluation. Values are taken from respective papers. For \cite{fuhl2016else}, we used the implementation provided by the authors with eye regions from facial feature alignment. * = value estimated from authors' graphs. Performance of {\textbf{\footnotesize REG-MR-G}} on GI4E is omitted since GI4E was used for training. The three best methods in each category are marked with \textcolor{ForestGreen}{green}, \textcolor{Orange}{orange}, and \textcolor{red}{red} respectively.}
	\label{tbl:quantitative_evaluation}
\end{table*}

We evaluate several versions of our method. To evaluate against alternative approaches and illustrate the effect of circle refinement we evaluate a regressor trained on manually annotated data with ({\textbf{\footnotesize REG-MR}}) and without ({\textbf{\footnotesize REG-M}}) circle refinement.  We also evaluate a regressor trained on automatically annotated data ({\textbf{\footnotesize REG-AR}}) and show how it compares to {\textbf{\footnotesize REG-MR}}, the hand crafted approach used to generate annotations ({\textbf{\footnotesize HC}}), and the competition. To evaluate the regressors trained on manual annotations we use the MPIIGaze dataset \cite{zhang15_cvpr} for training, which has $10229$ cropped out eye images with eye corners and center annotations. To test {\textbf{\footnotesize REG-AR}}, we need a dataset where the entire face is visible, thus we use the GI4E dataset with flipped images for training. Since GI4E is smaller than MPIIGaze, the regressor trained on it works marginally worse than the regressor trained on MPIIGaze, but nevertheless achieves state-of-the-art performance. We indicate the dataset used for training as a suffix to the method's name (-M for MPIIGaze and -G for GI4E). Figure \ref{fig:quantitative_evaluation_reg_man} left shows that for errors below $0.05$, for the two high-res datasets (GI4E and Talking Face), circle refinement leads to significant boost in accuracy. This is particularly apparent on the Talking Face dataset, where the accuracy for $e \le 0.025$ increased from $18.70\%$ to $65.78\%$. For the BioID dataset, foregoing refinement is marginally better. This is in line with our previous observation, and is likely due to the low resolution and poor quality of the images. Our method achieves state of the art performance across all the three datasets. In particular, for $e \le 0.05$ {\textbf{\footnotesize REG-MR-M}} ({\textbf{\footnotesize REG-M-M}}) achieves the accuracy of $95.07\%$ ($95.27\%$) on BioID, $99.27\%$ ($99.03\%$) on GI4E, and $95.68\%$ ($95.62\%$) on the Talking Face dataset. 

Recall that the evaluation is restricted to images where facial alignment passed verification. On GI4E and Talking Face datasets combined, only one image failed verification. However, on BioID $62$ images failed verification compared to only $6$ images where a face was not detected. Evaluating {\textbf{\footnotesize REG-MR-M}} on all images with a detected face yields a performance of $67.19\%$ for $e \le 0.025$, $94.19\%$ for $e \le 0.05$, $99.47\%$ for $e \le 0.1$, and $100\%$ for $e \le 0.25$, which is only marginally worse than the method with facial verification and still out-performs the competition. Future improvements to facial feature alignment will remove this gap in performance. Table \ref{tbl:quantitative_evaluation} summarizes the results.

Next, we compare the performance of the automatically trained regressor ({\textbf{\footnotesize REG-AR-G}}) to the hand-crafted approach that generated its annotations ({\textbf{\footnotesize HC}}), as well as to {\textbf{\footnotesize REG-MR-G}} trained on manually annotated GI4E data. The results are shown in Figure \ref{fig:quantitative_evaluation_reg_man} right and are included in Table \ref{tbl:quantitative_evaluation}. Observe that {\textbf{\footnotesize REG-AR-G}} outperforms the hand-crafted method on both the train (GI4E) and the test (BioID and Talking Face) sets. Moreover, on the test sets its performance is close to {\textbf{\footnotesize REG-MR-G}}.

\subsection{Qualitative Evaluation}
\begin{figure*}[t]
	\begin{center}
		\includegraphics[width=0.9\linewidth]{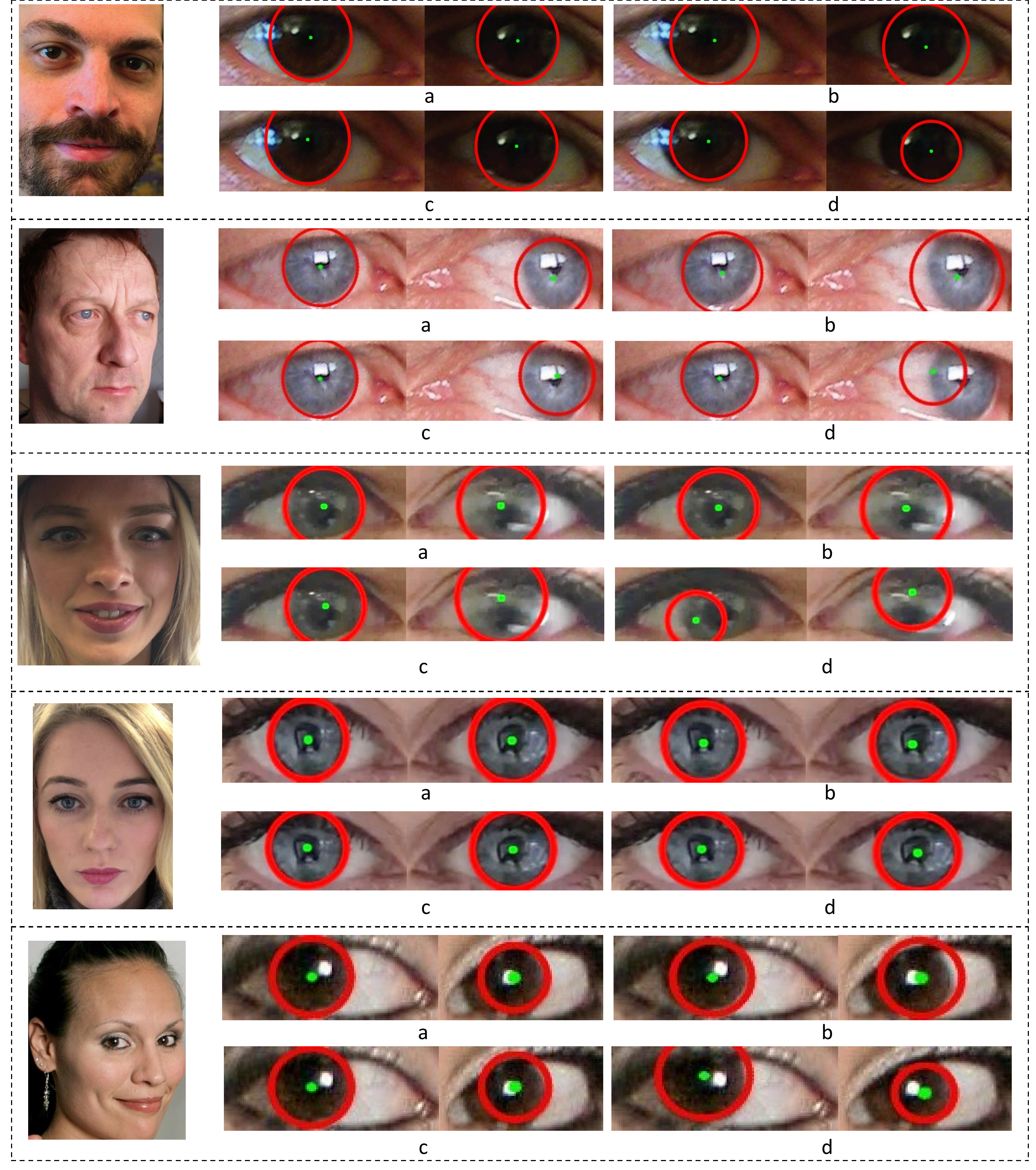}
	\end{center}
	\caption{Qualitative evaluation. For each image we show the results of (a) {\textbf{\footnotesize REG-MR-G}}, (b) {\textbf{\footnotesize REG-M-G}}, (c) {\textbf{\footnotesize REG-AR-G}}, and (d) {\textbf{\footnotesize HC}}.}
	\label{fig:qualitative_evaluation_selected}
\end{figure*}

For qualitative evaluation we compare the performance of {\textbf{\footnotesize REG-MR-G}}, {\textbf{\footnotesize REG-M-G}}, {\textbf{\footnotesize REG-AR-G}}, and {\textbf{\footnotesize HC}}. For consistency, all regressors were trained on GI4E. Figure \ref{fig:qualitative_evaluation_selected} shows selected results. The first four examples illustrate the increased accuracy when using circle refinement. Furthermore, the first three examples illustrate a failure of {\textbf{\footnotesize HC}} on at least one eye while {\textbf{\footnotesize REG-AR-G}}, which was trained using {\textbf{\footnotesize HC}}, is performing well. Finally, in the provided examples we observe no significant difference in quality between the manually and the automatically trained regressors, as well as their ability to cope with challenging imaging conditions such as poor lighting ($1$st and $3$rd images) and motion blur (left eye in $3$rd image).

One failure mode of our approach is when the pupils are near the eye corners. This is especially true for the inner corner (left eye in the last example of  Figure \ref{fig:qualitative_evaluation_selected}), where the proximity to the nose creates strong edges that likely confuse the HoG descriptor. The lack of training data with significant face rotation may also be a contributing factor. Training on more rotated faces, as well as using the skin color to downweigh the gradients from the nose, should help alleviate this issue.

\section{Conclusions}
\label{sec:Conclusions}
We presented a novel method for eye center localization. State-of-the-art performance is achieved by localizing the eyes using facial feature regression and then detecting eye centers using a cascade of regression forests with HoG features. Combining the regressor with a robust circle fitting step for iris refinement results in both robust and \textbf{accurate} localization. Using a hand-crafted eye center detector, the regressor can be trained automatically. The resulting regressor out-performs the hand-crafted method, works nearly as well as the manually trained regressor, and performs favorably compared to competing approaches. 

As a result of our ongoing work in this research area, we created an internal database consisting of $726$ very precisely labeled iris images. We observed that, although the number of photos were limited, for $e < 0.025$ we obtained better results using the methods outlined in this paper but trained on this highly-accurate database.  As part of our future work, we aim to expand this database and observe the extent to which the accuracy can be improved.

\bibliographystyle{ieee}
\bibliography{IrisDetection_2017}

\begin{thebibliography}{10}\itemsep=-1pt

\bibitem{bioid}
Bioid dataset.
\newblock \url{https://www.bioid.com/About/BioID-Face-Database}.

\bibitem{talkingface}
Talking face dataset.
\newblock
  \url{http://www-prima.inrialpes.fr/FGnet/data/01-TalkingFace/talking_face.html}.

\bibitem{ahuja2016eye}
K.~Ahuja, R.~Banerjee, S.~Nagar, K.~Dey, and F.~Barbhuiya.
\newblock Eye center localization and detection using radial mapping.
\newblock In {\em ICIP}, pages 3121--3125, 2016.

\bibitem{antonakos2015feature}
E.~Antonakos, J.~Alabort-i Medina, G.~Tzimiropoulos, and S.~P. Zafeiriou.
\newblock Feature-based lucas--kanade and active appearance models.
\newblock {\em IEEE Transactions on Image Processing}, 24(9):2617--2632, 2015.

\bibitem{gi4e}
M.~Ariz, J.~J. Bengoechea, A.~Villanueva, and R.~Cabeza.
\newblock A novel 2{D}/3{D} database with automatic face annotation for head
  tracking and pose estimation.
\newblock {\em Computer Vision and Image Understanding}, 148:201--210, 2016.

\bibitem{cao2014face}
X.~Cao, Y.~Wei, F.~Wen, and J.~Sun.
\newblock Face alignment by explicit shape regression.
\newblock {\em International Journal of Computer Vision}, 107(2):177--190,
  2014.

\bibitem{daugman2004iris}
J.~Daugman.
\newblock How iris recognition works.
\newblock {\em IEEE Transactions on circuits and systems for video technology},
  14(1):21--30, 2004.

\bibitem{dollar2009integral}
P.~Doll{\'a}r, Z.~Tu, P.~Perona, and S.~Belongie.
\newblock Integral channel features.
\newblock In {\em BMVC}, 2009.

\bibitem{fuhl2015excuse}
W.~Fuhl, T.~K{\"u}bler, K.~Sippel, W.~Rosenstiel, and E.~Kasneci.
\newblock Excuse: Robust pupil detection in real-world scenarios.
\newblock In {\em International Conference on Computer Analysis of Images and
  Patterns}, pages 39--51. Springer, 2015.

\bibitem{fuhl2016else}
W.~Fuhl, T.~C. Santini, T.~K{\"u}bler, and E.~Kasneci.
\newblock Else: Ellipse selection for robust pupil detection in real-world
  environments.
\newblock In {\em Proceedings of the Ninth Biennial ACM Symposium on Eye
  Tracking Research \& Applications}, pages 123--130. ACM, 2016.

\bibitem{george2016fast}
A.~George and A.~Routray.
\newblock Fast and accurate algorithm for eye localisation for gaze tracking in
  low-resolution images.
\newblock {\em IET Computer Vision}, 10(7):660--669, 2016.

\bibitem{hansen2010eye}
D.~W. Hansen and Q.~Ji.
\newblock In the eye of the beholder: A survey of models for eyes and gaze.
\newblock {\em IEEE transactions on pattern analysis and machine intelligence},
  32(3):478--500, 2010.

\bibitem{kawaguchi2000detection}
T.~Kawaguchi, D.~Hidaka, and M.~Rizon.
\newblock Detection of eyes from human faces by {H}ough transform and
  separability filter.
\newblock In {\em ICIP}, volume~1, pages 49--52, 2000.

\bibitem{kazemi2014one}
V.~Kazemi and J.~Sullivan.
\newblock One millisecond face alignment with an ensemble of regression trees.
\newblock In {\em Proceedings of the IEEE Conference on Computer Vision and
  Pattern Recognition}, pages 1867--1874, 2014.

\bibitem{levinshteinSIP2017}
A.~Levinshtein, E.~Phung, and P.~Aarabi.
\newblock Hybrid eye center localization using cascaded regression and robust
  circle fitting.
\newblock In {\em Global Conference on Signal and Information Processing}.
  IEEE, 2017.

\bibitem{li2005starburst}
D.~Li, D.~Winfield, and D.~J. Parkhurst.
\newblock Starburst: A hybrid algorithm for video-based eye tracking combining
  feature-based and model-based approaches.
\newblock In {\em CVPRW}, pages 79--79, 2005.

\bibitem{markuvs2014eye}
N.~Marku{\v{s}}, M.~Frljak, I.~S. Pand{\v{z}}i{\'c}, J.~Ahlberg, and
  R.~Forchheimer.
\newblock Eye pupil localization with an ensemble of randomized trees.
\newblock {\em Pattern recognition}, 47(2):578--587, 2014.

\bibitem{ren2014face}
S.~Ren, X.~Cao, Y.~Wei, and J.~Sun.
\newblock Face alignment at 3000 fps via regressing local binary features.
\newblock In {\em Proceedings of the IEEE Conference on Computer Vision and
  Pattern Recognition}, pages 1685--1692, 2014.

\bibitem{skodras2015precise}
E.~Skodras and N.~Fakotakis.
\newblock Precise localization of eye centers in low resolution color images.
\newblock {\em Image and Vision Computing}, 36:51--60, 2015.

\bibitem{swirski2012robust}
L.~{\'S}wirski, A.~Bulling, and N.~Dodgson.
\newblock Robust real-time pupil tracking in highly off-axis images.
\newblock In {\em Proceedings of the Symposium on Eye Tracking Research and
  Applications}, pages 173--176. ACM, 2012.

\bibitem{tian2016accurate}
D.~Tian, G.~He, J.~Wu, H.~Chen, and Y.~Jiang.
\newblock An accurate eye pupil localization approach based on adaptive
  gradient boosting decision tree.
\newblock In {\em Visual Communications and Image Processing (VCIP), 2016},
  pages 1--4. IEEE, 2016.

\bibitem{timm2011accurate}
F.~Timm and E.~Barth.
\newblock Accurate eye centre localisation by means of gradients.
\newblock {\em VISAPP}, 11:125--130, 2011.

\bibitem{toennies2002feasibility}
K.~Toennies, F.~Behrens, and M.~Aurnhammer.
\newblock Feasibility of hough-transform-based iris localisation for
  real-time-application.
\newblock In {\em Pattern Recognition, 2002. Proceedings. 16th International
  Conference on}, volume~2, pages 1053--1056. IEEE, 2002.

\bibitem{valenti2012accurate}
R.~Valenti and T.~Gevers.
\newblock Accurate eye center location through invariant isocentric patterns.
\newblock {\em IEEE transactions on pattern analysis and machine intelligence},
  34(9):1785--1798, 2012.

\bibitem{wood2014eyetab}
E.~Wood and A.~Bulling.
\newblock Eyetab: Model-based gaze estimation on unmodified tablet computers.
\newblock In {\em Proceedings of the Symposium on Eye Tracking Research and
  Applications}, pages 207--210. ACM, 2014.

\bibitem{xiong2013supervised}
X.~Xiong and F.~De~la Torre.
\newblock Supervised descent method and its applications to face alignment.
\newblock In {\em Proceedings of the IEEE conference on computer vision and
  pattern recognition}, pages 532--539, 2013.

\bibitem{zhang15_cvpr}
X.~Zhang, Y.~Sugano, M.~Fritz, and A.~Bulling.
\newblock Appearance-based gaze estimation in the wild.
\newblock In {\em Proc. of the IEEE International Conference on Computer Vision
  and Pattern Recognition (CVPR)}, pages 4511--4520, June 2015.

\bibitem{zhou2015precise}
M.~Zhou, X.~Wang, H.~Wang, J.~Heo, and D.~Nam.
\newblock Precise eye localization with improved sdm.
\newblock In {\em ICIP}, pages 4466--4470, 2015.

\end{thebibliography}

\end{document}